\documentclass{article}
\usepackage{spconf,amsmath,graphicx}
\usepackage{mathtools}
\usepackage{bbold}
\usepackage{algorithm} 
\usepackage{algpseudocode, setspace}
\usepackage{siunitx}
\usepackage[center]{subfigure}
\usepackage{graphicx,float}
\usepackage[table]{xcolor}
\definecolor{gray}{rgb}{0.85,0.85,0.85}
\usepackage{tabularray}
\usepackage{booktabs,arydshln}
\usepackage{xcolor}
\usepackage[export]{adjustbox}

\usepackage{hyperref}
\hypersetup{
	colorlinks   = true,    
	urlcolor     = blue,    
	linkcolor    = red,     
	citecolor    = green    
}

\graphicspath{{IMAGES/}}

\newcommand{\overbar}[1]{\mkern 1.5mu\overline{\mkern-1.5mu#1\mkern-1.5mu}\mkern 1.5mu}
\definecolor{dg}{rgb}{0.0, 0.5, 0.0}

\title{Fuzzy-conditioned diffusion and diffusion projection attention applied to facial image correction}
\name{Majed El Helou}
\address{Media Technology Center, ETH Z\"urich, Switzerland}

\begin{document}
\maketitle
\begin{abstract}
Image diffusion has recently shown remarkable performance in image synthesis and implicitly as an image prior. Such a prior has been used with conditioning to solve the inpainting problem, but only supporting binary user-based conditioning. 

We derive a fuzzy-conditioned diffusion, where implicit diffusion priors can be exploited with controllable strength. Our fuzzy conditioning can be applied pixel-wise, enabling the modification of different image components to varying degrees. Additionally, we propose an application to facial image correction, where we combine our fuzzy-conditioned diffusion with diffusion-derived attention maps. Our map estimates the degree of anomaly, and we obtain it by projecting on the diffusion space. We show how our approach also leads to interpretable and autonomous facial image correction.
\end{abstract}


\noindent\let\thefootnote\relax\footnotetext{\url{https://github.com/majedelhelou/FC-Diffusion}}

\section{Introduction} \label{sec:intro}
Denoising diffusion methods~\cite{sohl2015deep} are effective probabilistic image generation models~\cite{ho2020denoising} that learn to synthesize images by gradually recovering them from noisier distributions. They have good sample quality due to their pixel-wise operation, for instance compared with variational auto-encoders~\cite{pandey2021vaes,pandey2022diffusevae}, or optimal transport distribution mapping~\cite{daniels2021score}. Denoising diffusion currently remains an active research topic, with recent work on balancing the advantages of regular and residual latents~\cite{benny2022dynamic} and for balancing the trade-off between diversity and high photorealism~\cite{sehwag2022generating}. For a more extensive overview, we refer the reader to the survey by Croitoru~\textit{et al.}~\cite{croitoru2022diffusion}. 

As image generators, diffusion models implicitly learn image distribution representations that can be exploited in a multitude of applications~\cite{chung2022diffusion}, similar to adversarially trained priors~\cite{el2022bigprior}. For instance, their implicit prior is useful for image-to-image translation to map across different domains~\cite{sasaki2021unit,saharia2022palette}, for image super-resolution~\cite{saharia2022image}, as well as for synthesizing images with text-based conditioning~\cite{saharia2022photorealistic}. Their probabilistic nature also enables diverse image completion through randomized sampling~\cite{horita2022structure}. Stochastic differential equation models were recently used for image editing where the user can control the degree of projection on the perturbed space~\cite{meng2021sdedit}. A direct inpainting application is presented in~\cite{lugmayr2022repaint}, providing both state-of-the-art results as well as different diverse outputs due to the stochasticity of the diffusion prior. In contrast, we propose a generalization of inpainting to the non-binary setting, in an effort towards prior contribution control~\cite{el2020blind,el2022bigprior} and control over network hallucination~\cite{el2021deep}. 

We present a fuzzy-conditioned diffusion where inpainting is a special case. Our model can synthesize novel images from different stochastic samples, while conditioning on a given input image in an adjustable way. The conditioning can be spatially varying and can also be fuzzy (non-binary) in nature. In other words, the diffusion model can be conditioned on an input image to a varying degree, and this degree can change across the image, thus exploiting the underlying pixel-based formulation of denoising diffusion. 

Diffusion models can also be used for out-of-distribution detection~\cite{liu2022diffusion,wyatt2022anoddpm,graham2022denoising}. The first challenge in out-of-distribution detection is to learn a good image prior for the in-distribution data, which diffusion models can effectively do. Another challenge is to then classify out-of-distribution test cases. The approach in~\cite{graham2022denoising}, inspired by previous work using deep auto-encoders~\cite{gong2019memorizing}, classifies test cases based on the full stack of reconstruction errors across a sequence of projections on all latent spaces of a diffusion model. A second approach proposes a combination of a diffusion reconstruction with a learned discriminator network to obtain the final classification~\cite{liu2022diffusion}. Another more medically-oriented approach aims to detect and quantify deviations on the pixel-wise level with a non-Gaussian diffusion~\cite{wyatt2022anoddpm}. Building on these methods, we construct a diffusion projection attention map on a standard denoising diffusion model. We then exploit this map in our fuzzy-conditioned diffusion to guide an autonomous image correction on a facial image application. 

We summarize our contributions as follows. We derive a novel approach to condition diffusion models on images, such that the conditioning can be spatially varying as well as fuzzy in strength. This enables both a generalization of inpainting where the degree of hallucination is controlled, and a generalization of image conditioning to be spatially adaptive. We additionally provide a projection-based anomaly map, enabling our model to perform autonomous face correction. 


\section{Proposed method}
We exploit diffusion models~\cite{sohl2015deep} to propose a diffusion with a fuzzy image conditioning (Sec.~\ref{sec:fuzzy}) and to obtain a guide-free attention map through projection (Sec.~\ref{sec:projection}), the combination of which is applied for autonomous facial image correction.

\subsection{Mathematical formulation} \label{sec:ddpm}
We begin with a brief overview of DDPM~\cite{ho2020denoising}, which learns to map a Gaussian distribution to a data distribution, similar to other generative methods. Assuming the image data distribution to be given by $x_0\sim q(x_0)$, we follow a Markovian noising process to gradually add Gaussian noise to the data
\begin{equation}
    q(x_t|x_{t-1}) \coloneqq \mathcal{N}(x_t; \sqrt{1-\beta_t} x_{t-1}, \beta_t \mathbb{1}),
\end{equation}
where $\beta_t$ dictates a variance schedule across iterations.
We can sample in the diffusion latent space $x_t \sim q(x_t|x_0)$ directly with a single procedure from the distribution
\begin{equation} \label{eq:sampling}
    q(x_t|x_0) = \mathcal{N}(x_t; \sqrt{\overbar{\alpha}_t} x_0, (1-\overbar{\alpha}_t) \mathbb{1}),
\end{equation}
where $\alpha_t\coloneqq1-\beta_t$ and $\overbar{\alpha}_t \coloneqq \prod_{d=0}^{t} \alpha_d$. This leads to $x_t = \sqrt{\overbar{\alpha}_t} x_0 + \epsilon \sqrt{1-\overbar{\alpha}_t}, \epsilon \sim \mathcal{N}(0,\mathbb{1})$, where $\epsilon$ is a sampled noise tensor of same dimensions as $x_0$. The conditional posterior distribution is derived with Bayes' theorem as
\begin{equation}
    q(x_{t-1}|x_t,x_0) = \mathcal{N}(x_{t-1}; \tilde{\mu}(x_t,x_0), \tilde{\beta}_t\mathbb{1}),
\end{equation}
where the Gaussian parameters are respectively given by
\begin{equation}
    \begin{split}
        & \tilde{\mu}(x_t,x_0) \coloneqq \frac{\sqrt{\overbar{\alpha}_{t-1}}\beta_t}{1-\overbar{\alpha}_t}x_0 +
        \frac{\sqrt{\alpha}_t(1-\overbar{\alpha}_{t-1})}{1-\overbar{\alpha}_t}x_t, \\
        & \tilde{\beta}_t \coloneqq \frac{1-\overbar{\alpha}_{t-1}}{1-\overbar{\alpha}_t} \beta_t.
    \end{split}
\end{equation}
The true $q(x_{t-1}|x_t)$ is estimated with a learned distribution 
\begin{equation}
    p_\theta(x_{t-1}|x_t) \coloneqq \mathcal{N}(x_{t-1}; \mu_\theta(x_t,t), \Sigma_\theta(x_t,t)),
\end{equation}
where the mean $\mu_\theta$ and the covariance matrix $\Sigma_\theta$ are learned and predicted by a $\theta$-parameterized network, and the covariance matrix can be fixed to a constant in practice~\cite{ho2020denoising}. Learning to predict $\tilde{\mu}$ amounts to predicting $x_0$, or equivalently $\epsilon$ due to the conditioning on $x_t$ in the posterior, leading to a learning objective of minimizing ${E}_{t,x_0\sim q(x_0),\epsilon \sim \mathcal{N}(0,\mathbb{1})} [||\epsilon-\epsilon_{\theta}(x_t,t)||^2]$, where $\epsilon_{\theta}$ is predicted by a neural network and the norm being a pixel-wise $\ell_2$ norm.

\subsection{Fuzzy-conditioned image diffusion} \label{sec:fuzzy}
To condition the diffusion, the synthetic image $x_{t-1}$ and the projected input image $x^r_{t-1}$ (re-noised conditioning input) can be fused with a weight map $m$ throughout the diffusion process. The projection $x^r_{t-1}$ is sampled with Eq.~\eqref{eq:sampling} at $t-1$, and $x_{t-1}$ is obtained from our modified diffusion process. Unlike inpainting, our map is fuzzy rather than binary and the fusion step requires distribution matching such that each resulting pixel corresponds to the correct Markovian $q(x_t|x_{t-1})$ distribution. 
The next forward step to obtain $x_t$ is preceded by the following transformation over $x_{t-1}$
\begin{equation}
    x_m \coloneqq \sqrt{\overbar{\alpha}_{t-1}} x
    + \frac{
    m \cdot x^r_{t-1} + (1-m)\cdot x_{t-1} - \sqrt{\overbar{\alpha}_{t-1}} x
    }{
    \sqrt{1-2\cdot m + 2\cdot m^2}
    },
\end{equation}
using a fuzzy weight map $m$ that can be obtained from Sec.~\ref{sec:projection} or can be any map of same dimensions as $x$ such as its values are in $[0,1]$. $x$ is the conditioning input image that becomes the ground-truth target when $m\to 1$ and gets ignored when $m\to 0$. The standardization coefficient $\sqrt{1-2\cdot m + 2\cdot m^2}$ comes from the fact that the resulting variance at $t-1$  of $x_m$ is given by $m^2 V\{x^r_{t-1}\} + (1-m)^2 V\{x_{t-1}\}$ where both variances $V\{x^r_{t-1}\}$ and $V\{x_{t-1}\}$ are made equivalent by design, matching $(1-\overbar{\alpha}_t)$. We then obtain $x_t$ by a forward step over $x_m$ instead of over $x_{t-1}$. Lastly, we use the harmonization resampling technique presented in~\cite{lugmayr2022repaint}, to promote high-level content consistency. The harmonization resampling consists of incorporating backward steps $J$ in the forward diffusion process, to bring information from future more refined steps back to earlier steps. This is achieved by resampling $x_t \sim \mathcal{N}(\sqrt{1-\beta_t} x_{t-1}, \beta_t \mathbb{1})$ for a fixed number of steps, and restarting the diffusion from that iteration. We summarize our method in Algorithm~\ref{algo}.

\begin{algorithm}[t]
	\caption{Our fuzzy-conditioned image diffusion} 
	\begin{algorithmic}[1]
    \setstretch{1.35}
        \State $x_T \sim \mathcal{N}(0,\mathbb{1})$
		\For {$t=T,\ldots, 1$}
			\For {$j=1,\ldots,J$}
				\State $\epsilon \sim \mathcal{N}(0,\mathbb{1})$ if $t>1$ else $\epsilon=0$
                \State $x^r_{t-1} = \sqrt{\overbar{\alpha}_{t-1}} x + \epsilon \sqrt{1-\overbar{\alpha}_{t-1}}$
                \State $\epsilon_2\sim \mathcal{N}(0,\mathbb{1})$ if $t>1$ else $\epsilon_2=0$
                \State $x_{t-1} = \frac{1}{\sqrt{\overbar{\alpha}_t}} (x_t-\frac{1-\alpha_t}{\sqrt{1-\overbar{\alpha}_t}}) \epsilon_\theta(x_t,t) + \tilde{\beta}_t*\epsilon_2$
                \State $x_m = m \cdot x^r_{t-1} + (1-m)\cdot x_{t-1} - \sqrt{\overbar{\alpha}_{t-1}} x$
                \State $x_m = \sqrt{\overbar{\alpha}_{t-1}} x + x_m / \sqrt{1-2\cdot m + 2\cdot m^2}$
	            \If {$j<J$ and $t>1$} {\newline
                    \hspace*{4.2em} $x_t\sim \mathcal{N}(\sqrt{1-\beta_{t}}x_{t-1}, \beta_{t}\mathbb{1})$
                } \EndIf
            \EndFor
		\EndFor \\
    \Return $x_0$
	\end{algorithmic} 
 \label{algo}
\end{algorithm}

\begin{figure*}
    \centering
    \SetTblrInner{hspan=minimal}
    \begin{tblr}{Q[valign=m,halign=m]Q[valign=m,halign=m]Q[valign=m,halign=m]Q[valign=m,halign=m]Q[valign=m,halign=m]Q[valign=m,halign=m]Q[valign=m,halign=m]Q[valign=m,halign=m]Q[valign=m,halign=m]}

      & \SetCell{green!10} \textcolor{dg}{Masked input} & \SetCell{green!10} \textcolor{dg}{$m_5$ (inpaint)} & \SetCell{green!10} \textcolor{dg}{$m_4$} & \SetCell{green!10} \textcolor{dg}{$m_3$} & \SetCell{green!10} \textcolor{dg}{$m_2$} & \SetCell{green!10} \textcolor{dg}{$m_1$} & \SetCell{green!10} \textcolor{dg}{$m_0$ (random)} \\

     \SetCell{gray} \rotatebox{90}{\parbox{2cm}{\centering Ours}} & 
     \SetCell{green!10} \includegraphics[width=0.11\textwidth, margin=0]{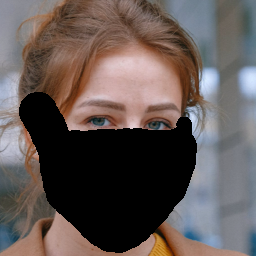} & 
     \SetCell{green!10} \includegraphics[width=0.11\textwidth]{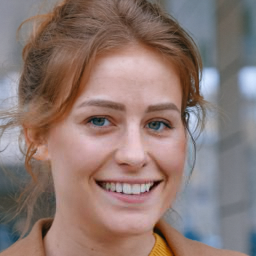} & 
     \SetCell{green!10} \includegraphics[width=0.11\textwidth]{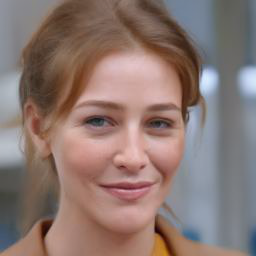} & 
     \SetCell{green!10} \includegraphics[width=0.11\textwidth]{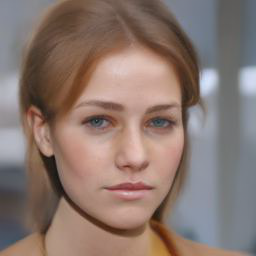} & 
     \SetCell{green!10} \includegraphics[width=0.11\textwidth]{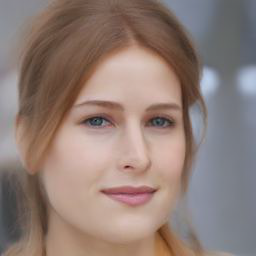} & 
     \SetCell{green!10} \includegraphics[width=0.11\textwidth]{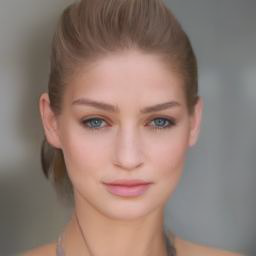} & 
     \SetCell{green!10} \includegraphics[width=0.11\textwidth]{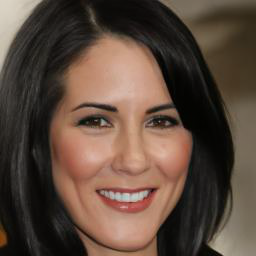} \\

     \SetCell{gray} \rotatebox{90}{\parbox{2cm}{\centering RePaint~\cite{lugmayr2022repaint}}} & 
     \SetCell{green!10} \includegraphics[width=0.11\textwidth, margin=0]{IMAGES/figFuzzy/masked.png} & 
     \SetCell{green!10} \includegraphics[width=0.11\textwidth]{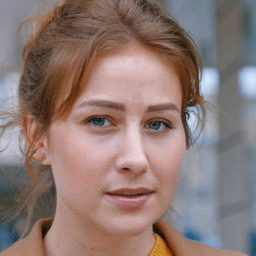} & \SetCell{green!10} \includegraphics[width=0.11\textwidth]{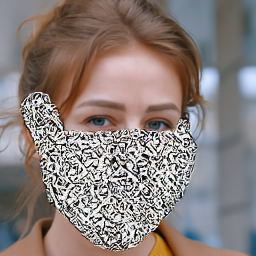} & 
     \SetCell{green!10} \includegraphics[width=0.11\textwidth]{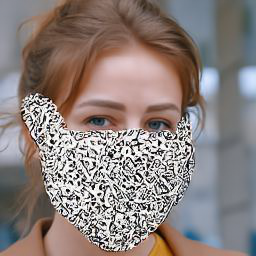} & 
     \SetCell{green!10} \includegraphics[width=0.11\textwidth]{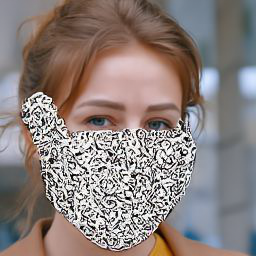} & 
     \SetCell{green!10} \includegraphics[width=0.11\textwidth]{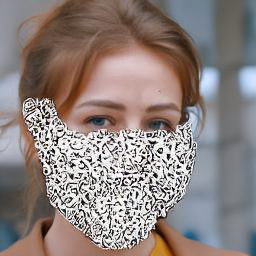} & 
     \SetCell{green!10} \includegraphics[width=0.11\textwidth]{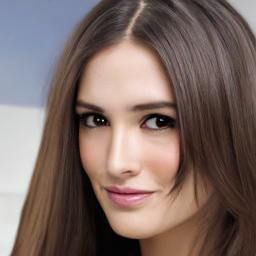} \\ 
     \SetCell{gray}\hline[dashed]
     
     \SetCell{gray} \rotatebox{90}{\parbox{2cm}{\centering Ours}} & 
     \SetCell{blue!10} \includegraphics[width=0.11\textwidth]{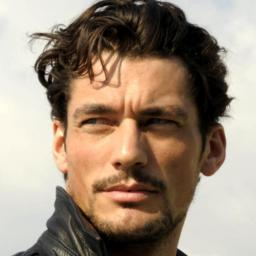} & 
     \SetCell{blue!10} \includegraphics[width=0.11\textwidth]{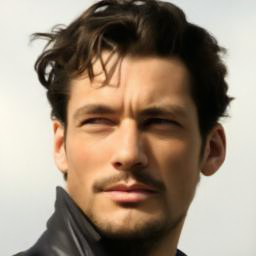} & 
     \SetCell{blue!10} \includegraphics[width=0.11\textwidth]{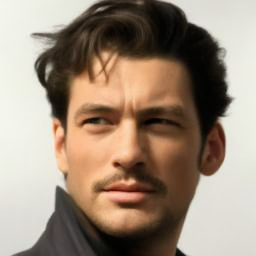} & 
     \SetCell{blue!10} \includegraphics[width=0.11\textwidth]{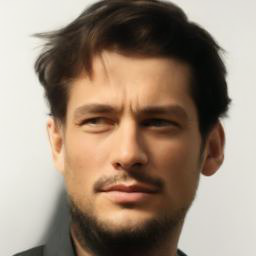} & 
     \SetCell{blue!10} \includegraphics[width=0.11\textwidth]{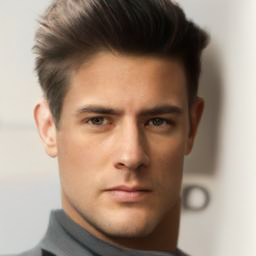} & 
     \SetCell{blue!10} \includegraphics[width=0.11\textwidth]{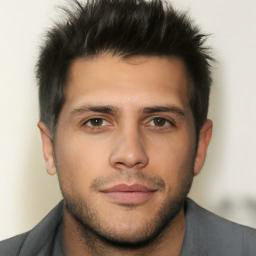} & 
     \SetCell{blue!10} \includegraphics[width=0.11\textwidth]{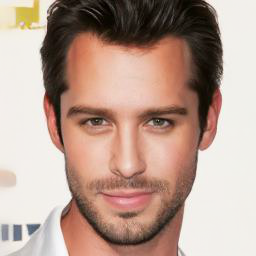} \\

     \SetCell{gray} \rotatebox{90}{\parbox{2cm}{\centering Ours}} & 
     \SetCell{blue!10} \includegraphics[width=0.11\textwidth]{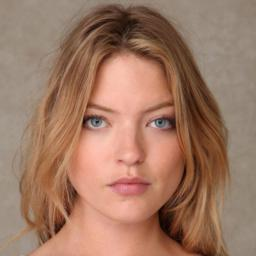} & 
     \SetCell{blue!10} \includegraphics[width=0.11\textwidth]{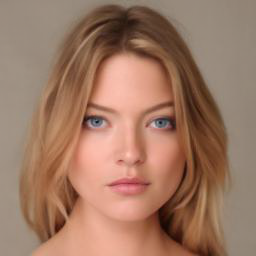} & 
     \SetCell{blue!10} \includegraphics[width=0.11\textwidth]{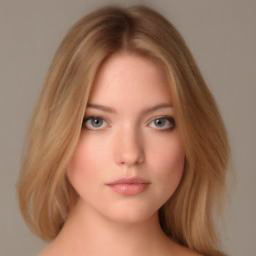} & 
     \SetCell{blue!10} \includegraphics[width=0.11\textwidth]{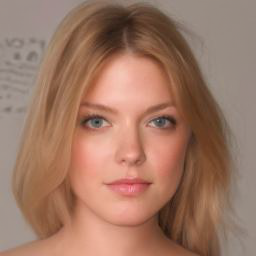} & 
     \SetCell{blue!10} \includegraphics[width=0.11\textwidth]{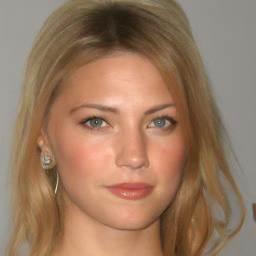} & 
     \SetCell{blue!10} \includegraphics[width=0.11\textwidth]{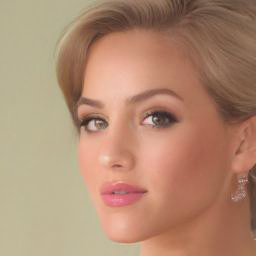} & 
     \SetCell{blue!10} \includegraphics[width=0.11\textwidth]{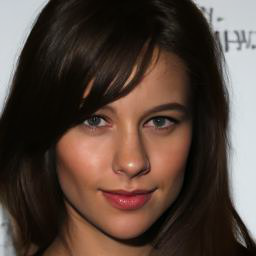} \\

    \end{tblr}  
    \vspace{-0.05cm}
    \caption{\textbf{Top:} Input with known mask, and generated image samples with uniform maps $m_i>m_j$ ($\forall i>j$) each multiplied by the inpainting mask. $m_5=1$ is standard inpainting, $m_0=0$ is an image synthesis, and in-between columns inpaint with fuzzy conditioning. We see that our approach enables varying the conditioning strength outside the mask, when a regular diffusion inpainting~\cite{lugmayr2022repaint} collapses. \textbf{Bottom:} our diffusion synthesis with spatially uniform and gradually decreasing input conditioning.}
    \vspace{-0.2cm}
    \label{fig:fuzzy}
\end{figure*}


\subsection{Diffusion projection attention} \label{sec:projection}
We exploit a diffusion network to obtain an inference-time attention map that we describe in this section. We project an image $x$ onto the latent space of the diffusion model following Eq.~\eqref{eq:sampling} $\forall t \in PS$, where $PS$ is a set of projection sampling iterations. We reconstruct the $\widehat{x}_t$ images with the diffusion model, and study the discrepancies between the input image and its reconstructions across all iterations. In particular, we consider the deviations in the discrepancies relative to standard discrepancies that can be expected from data within $q(x_0)$. The projection attention map we compute is given by
\begin{equation} \label{eq:map}
    A(x) \coloneqq \frac{1}{N}\sum_{t \in PS} \mathcal{H}\left( \frac{
    \phi(x-\widehat{x}_t ) -  \mu_t\{\mathcal{V}\}
    }{
    \sigma_t\{\mathcal{V}\}
    } \right),
\end{equation}
where $x$ is the input image (with potential anomaly or degradation), $\widehat{x}_t$ is the full forward-diffusion reconstruction of $x$ after $t$ sampling iterations, $\phi(\cdot)$ denotes a discrepancy measure which we set to the $\ell_2$-norm in our experiments, $\mathcal{H}$ is a truncation function that limits the normalized maps to the range $[\sigma, 6\sigma]$ such that $\mathcal{H}(x)=min(max(x,1),6)$, and $(\mu_t\{\mathcal{V}\}, \sigma_t\{\mathcal{V}\})$ are respectively defined as
\begin{equation}
    \begin{split}
    & \mu_t\{\mathcal{V}\} \coloneqq \frac{1}{|\mathcal{V}|} \sum_{z\in \mathcal{V}} \phi(z - \widehat{z}_t) , \\
    & \sigma_t\{\mathcal{V}\} \coloneqq \sqrt{
    \frac{1}{|\mathcal{V}|}\sum_{z\in \mathcal{V}}{
    ( \phi(z - \widehat{z}_t) - \mu_t\{\mathcal{V}\})^2}},
    \end{split}
\end{equation}
where $\mathcal{V}$ is a validation set corresponding to the training data of the diffusion model, and $\widehat{z}_t$ is defined in an analogous way to $\widehat{x}_t$ above, with the same diffusion projection approach. All operations are executed following the diffusion model's formulation in a pixel-wise manner.
For all experiments, $PS$ is the set given by $\{300,400,500,600\}$ and the final map $m$ is given by $m\coloneqq (1-A(x))^2$ to promote anomaly removal and preserve regions that are within one standard deviation.


\section{Experimental Evaluation}
\subsection{Experimental setup}
We experimentally evaluate each of our contributions separately, and our proposed application framework. In our application, we exploit fuzzy-conditioned image diffusion, combined with diffusion projection attention as a guiding map, for autonomous face correction. Given an input facial image, we autonomously detect anomalies and correct them using our weighted diffusion prior. All experiments use DDPM models trained on CelebA~\cite{liu2015deep} data with $256\times 256$ image resolution. 

\subsection{Fuzzy-conditioned image diffusion}
We illustrate how fuzzy-conditioned diffusion enables the generation of novel images with varying degrees of conditioning. We show in the top of Fig.~\ref{fig:fuzzy} a gradually decreasing degree of conditioning from left to right, coupled with an inpainting area (0 conditioning). Each column can be sampled practically \textit{infinitely many times}. RePaint~\cite{lugmayr2022repaint}, which cannot support fuzzy conditioning, fails on non-binary maps due to diffusion probability distribution mismatch. The bottom of Fig.~\ref{fig:fuzzy} shows synthesis examples with a uniform and gradually decreasing conditioning weight on the input.

\subsection{Diffusion projection attention}
\begin{figure}{}
    \centering
    \subfigure[$\mu_{300}\{\mathcal{V}\}$]{
        \includegraphics[width=0.11\textwidth]{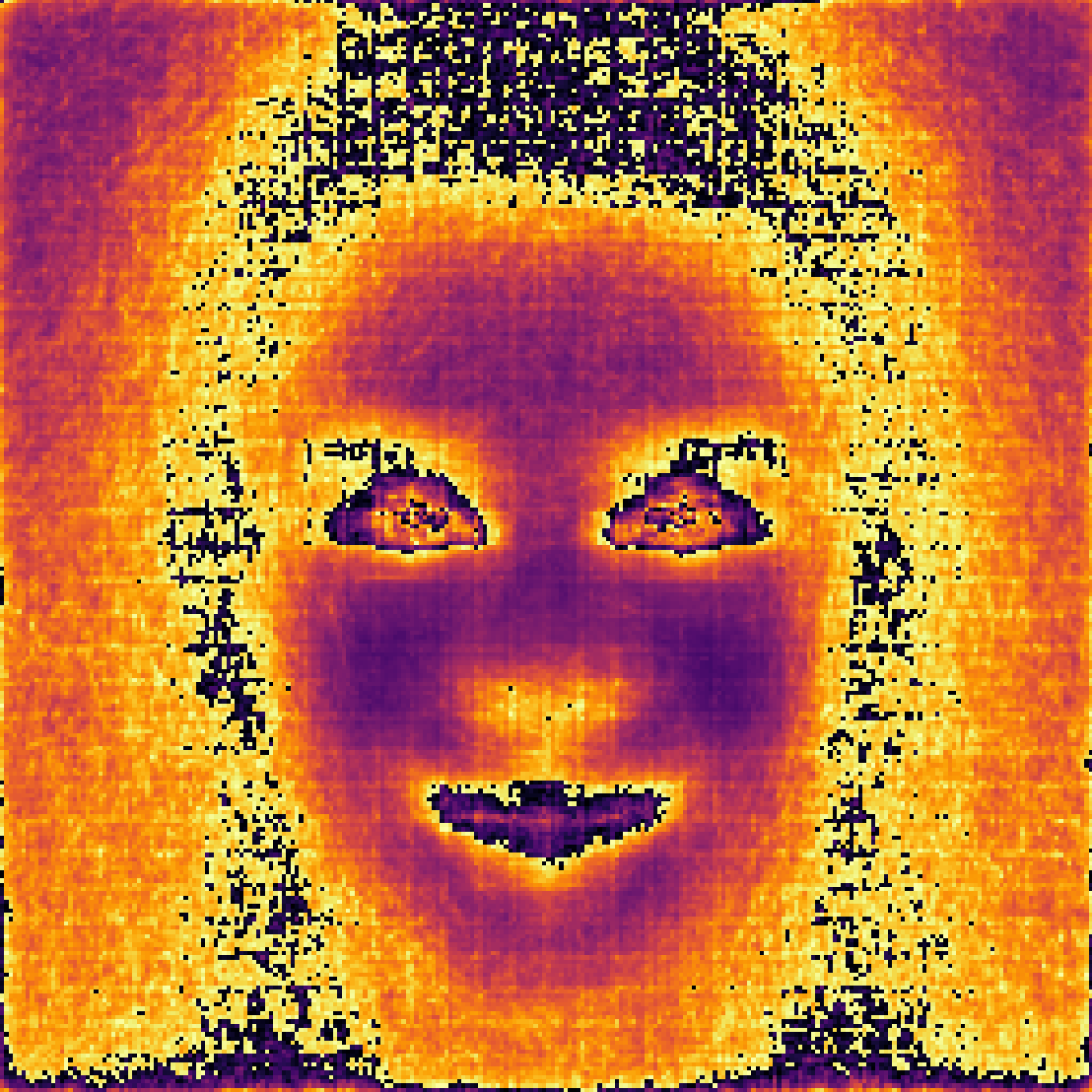}}
    \subfigure[$\mu_{400}\{\mathcal{V}\}$]{
        \includegraphics[width=0.11\textwidth]{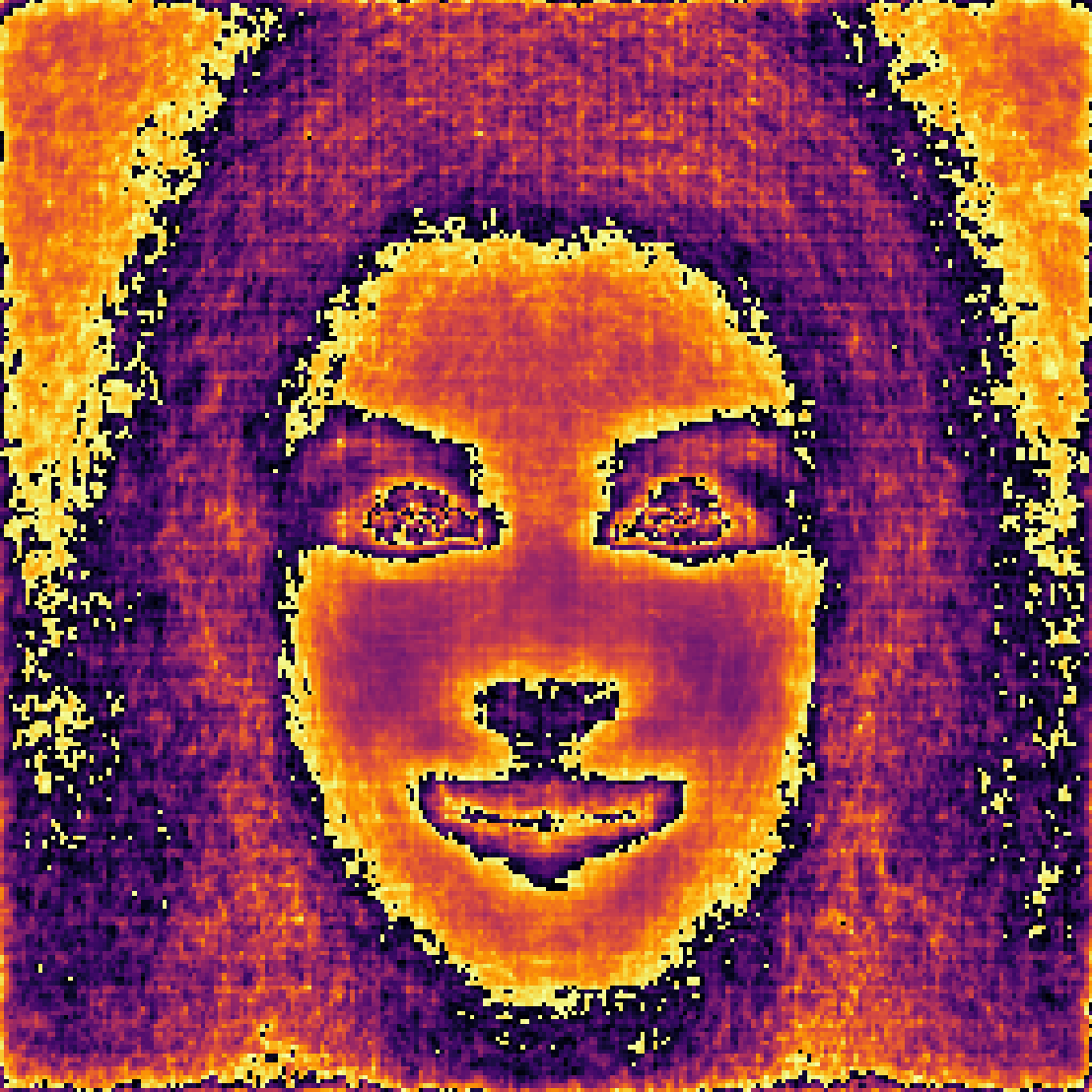}}
    \subfigure[$\mu_{500}\{\mathcal{V}\}$]{
        \includegraphics[width=0.11\textwidth]{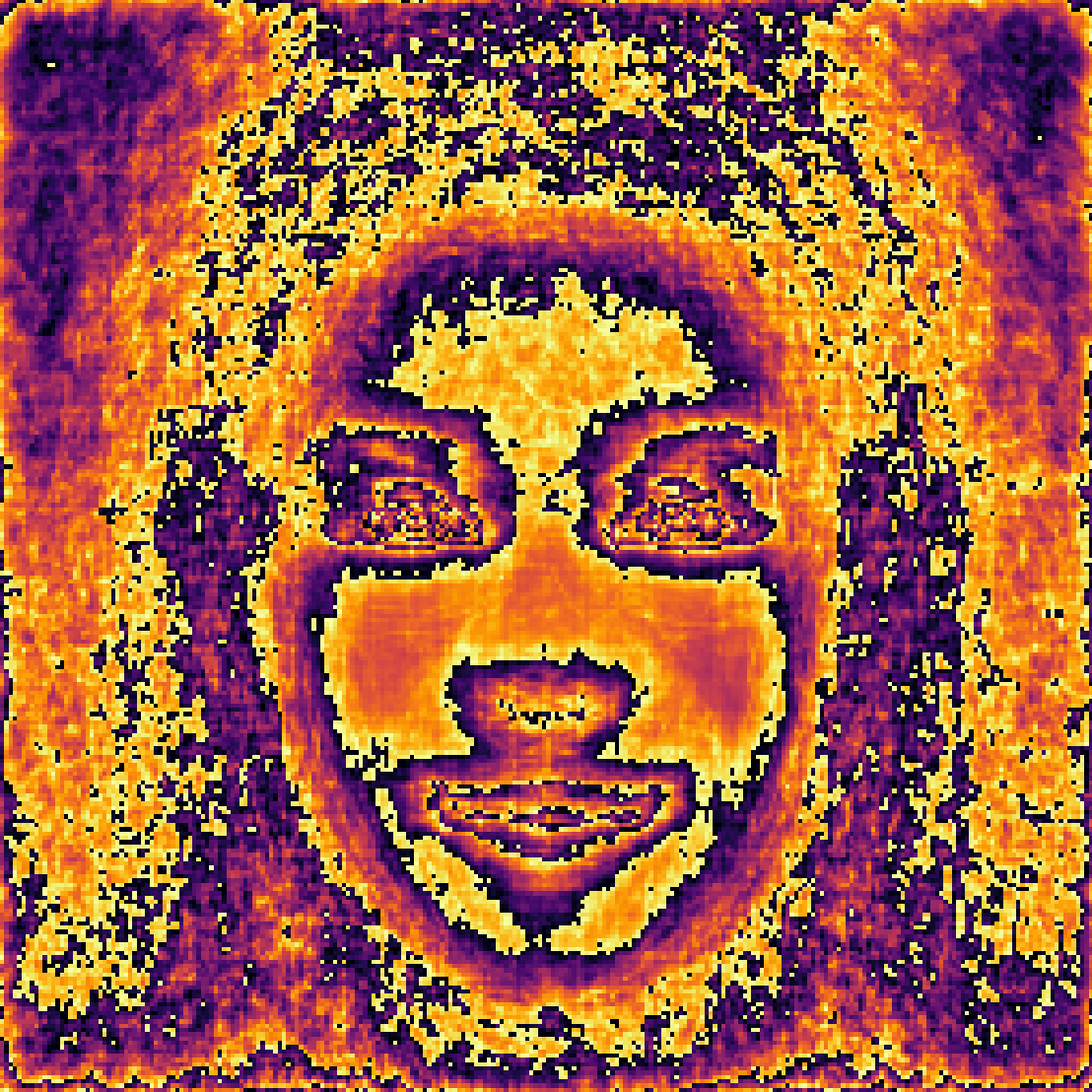}}
    \subfigure[$\mu_{600}\{\mathcal{V}\}$]{
        \includegraphics[width=0.11\textwidth]{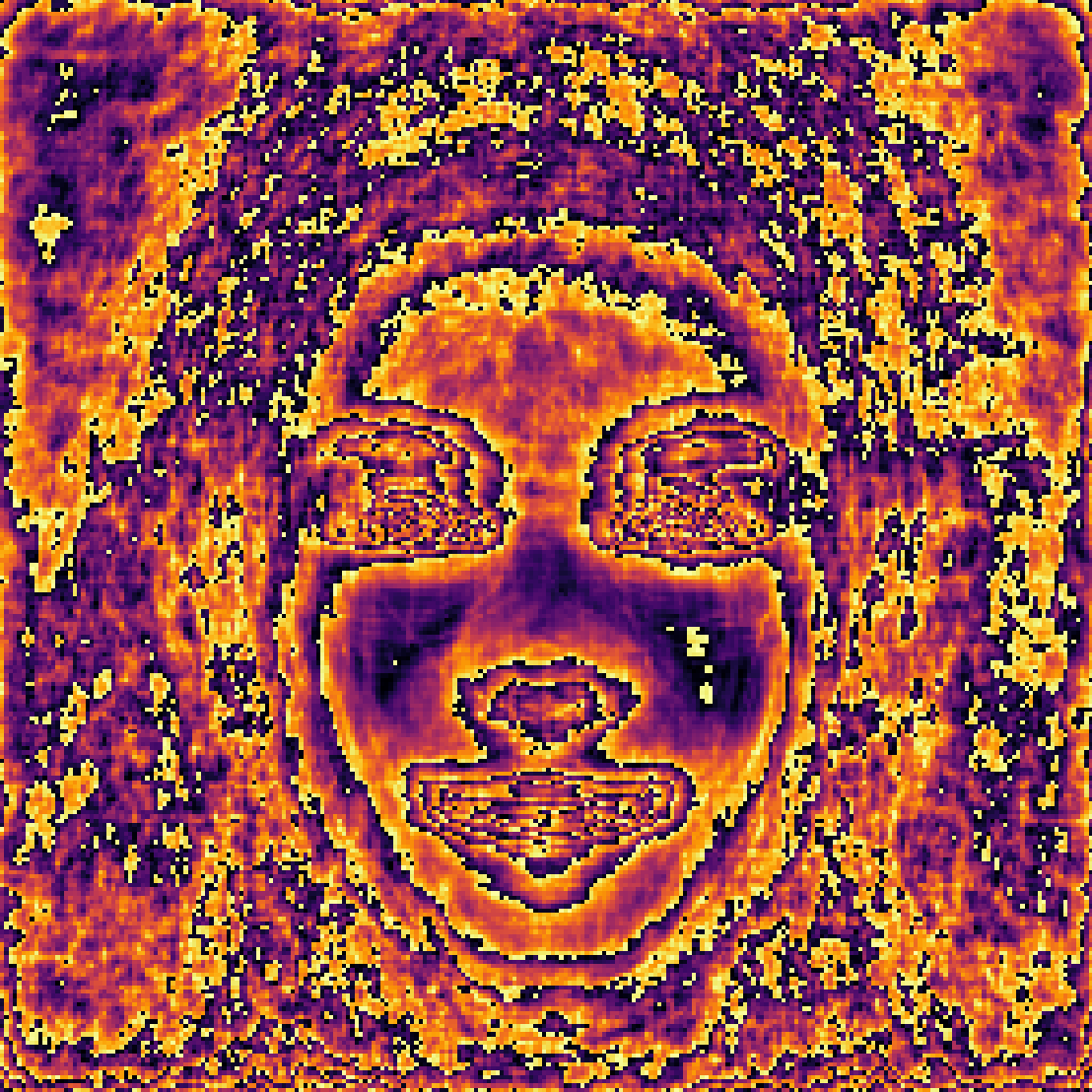}}

    \subfigure[Input $x$]{
        \includegraphics[width=0.11\textwidth]{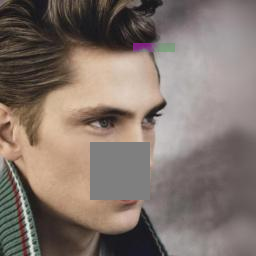}}
    \subfigure[$\phi(x-\widehat{x}_{\color{red}{300}})$]{
        \includegraphics[width=0.11\textwidth]{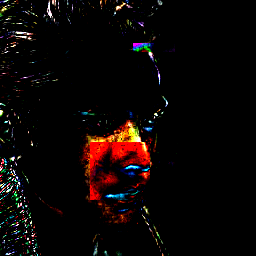}}
    \subfigure[$\phi(x-\widehat{x}_{\color{red}{400}})$]{
        \includegraphics[width=0.11\textwidth]{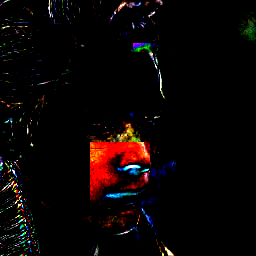}}
    \subfigure[$m(x)^*$]{
        \frame{\includegraphics[width=0.11\textwidth]{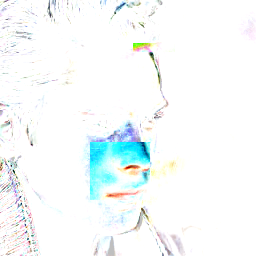}}}

    \subfigure[Input $x$]{
        \includegraphics[width=0.11\textwidth]{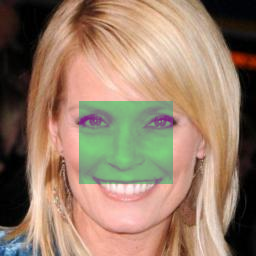}}
    \subfigure[$\phi(x-\widehat{x}_{\color{red}{500}})$]{
        \includegraphics[width=0.11\textwidth]{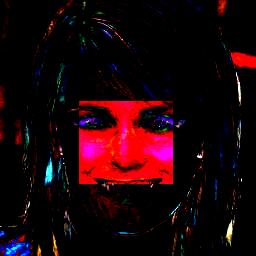}}
    \subfigure[$\phi(x-\widehat{x}_{\color{red}{600}})$]{
        \includegraphics[width=0.11\textwidth]{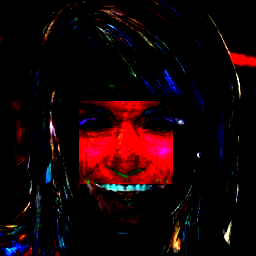}}
    \subfigure[$m(x)^*$]{
        \frame{\includegraphics[width=0.11\textwidth]{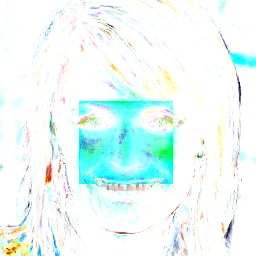}}}

    \vspace{-0.2cm}
    \caption{\textbf{Top row:} Expected value of $\phi(z-\widehat{z}_t)$ estimated over $\mathcal{V}$, for increasing diffusion projection depths $t$. \textbf{Middle and bottom rows:} Respectively a degraded input $x$, normalized projection deviations $\phi(x-\widehat{x}_t)$, and the anomaly map $m(x)$ ($^*$shown without scale adjustment for better visualization).}
    \label{fig:zscores}
\end{figure}

The top row of Fig.~\ref{fig:zscores} shows the mean maps at different diffusion projection depths. The maps are computed with $|\mathcal{V}|=1000$, which is empirically sufficient for stable convergence. The middle and bottom rows show two example images that contain anomalous pixel-dependent perturbations (we add a different offset on each pixel to reach a random threshold, within a randomly selected rectangle). For each example, we show the normalized projection deviations at different projection depths, as well as the overall map. We note how the map is able to autonomously detect the perturbations, without any prior knowledge of the test-time degradation model.

\subsection{Diffusion fuzzy-guided diffusion}
\begin{figure}[t]{}
    \centering
    \subfigure{
        \includegraphics[width=0.11\textwidth]{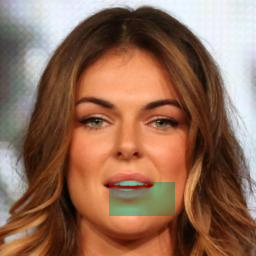}}
    \subfigure{
        \includegraphics[width=0.11\textwidth]{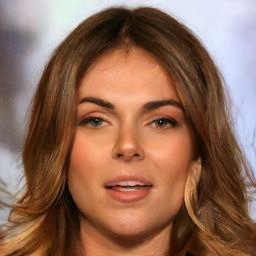}}
    \subfigure{
        \includegraphics[width=0.11\textwidth]{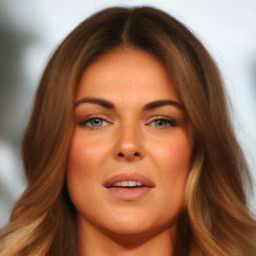}}
    \subfigure{
        \frame{\includegraphics[width=0.11\textwidth]{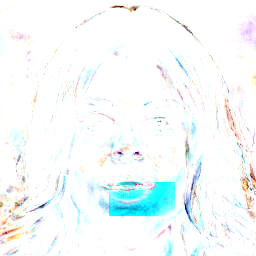}}}

    \subfigure{
        \includegraphics[width=0.11\textwidth]{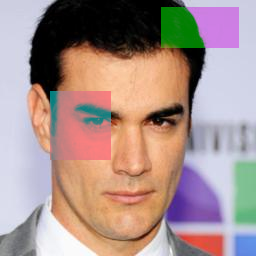}}
    \subfigure{
        \includegraphics[width=0.11\textwidth]{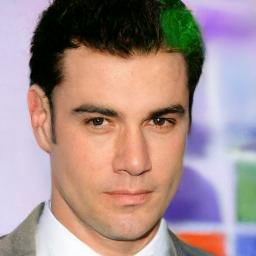}}
    \subfigure{
        \includegraphics[width=0.11\textwidth]{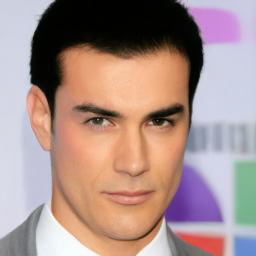}}
    \subfigure{
        \frame{\includegraphics[width=0.11\textwidth]{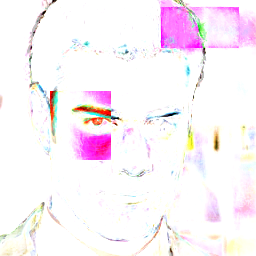}}}


    \subfigure{
        \includegraphics[width=0.11\textwidth]{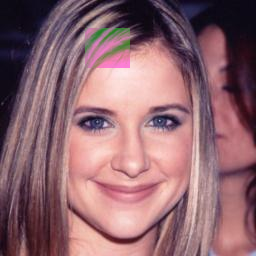}}
    \subfigure{
        \includegraphics[width=0.11\textwidth]{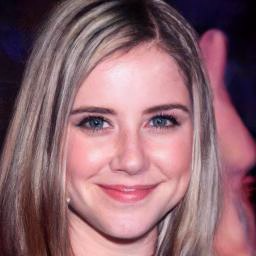}}
    \subfigure{
        \includegraphics[width=0.11\textwidth]{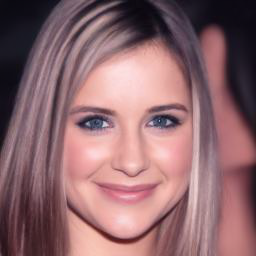}}
    \subfigure{
        \frame{\includegraphics[width=0.11\textwidth]{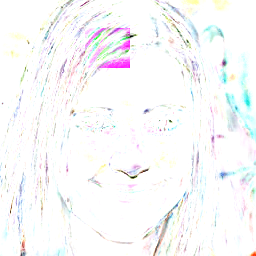}}}

    \addtocounter{subfigure}{-12}

    \subfigure[Input $x$]{
        \includegraphics[width=0.11\textwidth]{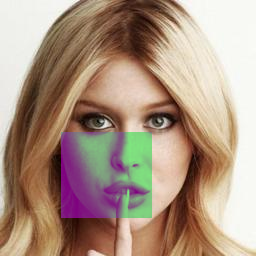}}
    \subfigure[DDPM~\cite{ho2020denoising}]{
        \includegraphics[width=0.11\textwidth]{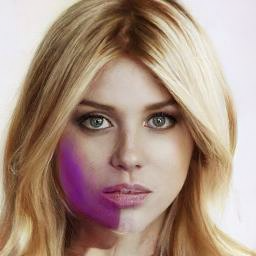}}
    \subfigure[Ours]{
        \includegraphics[width=0.11\textwidth,cfbox=blue 0.5pt 0pt]{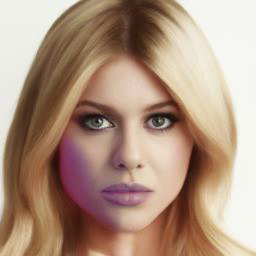}}
    \subfigure[Our $m(x)^*$]{
        \frame{\includegraphics[width=0.11\textwidth]{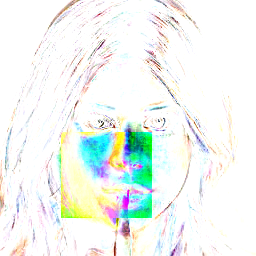}}}




    \vspace{-0.2cm}        
    \caption{\textbf{From left to right:} A degraded input image $x$, the restored output with $400$ iterations of DDPM~\cite{ho2020denoising}, our restored output and our human-interpretable fuzzy correction map $m(x)$ ($^*$shown with the same visualization as Fig.~\ref{fig:zscores}).}
    \label{fig:dgd_results}
\end{figure}

We apply a similar degradation algorithm as in the previous section, compute our conditioning map $m(x)$, and use it as a guide for our fuzzy-conditioned diffusion. Sample results are shown in Fig.~\ref{fig:dgd_results}, compared against DDPM projection~\cite{ho2020denoising} that is also a fully degradation-agnostic image correction. We can see that our results correct the degradations that overlap with the face (second row) while remaining more visually faithful to the input images. Furthermore, our final outputs are more human interpretable, as the guidance map illustrates the degree of hallucination per pixel. We lastly note that our method is conditioned on a facial image prior and, as such, is prone to treating non-facial occlusions as degradations to be removed (see the finger removal in the bottom row, blue frame).

\section{Conclusion} \label{sec:ccl}
We derive a novel fuzzy-conditioned image diffusion capable of generating random images with pixel-wise varying degrees of image conditioning. We also present a diffusion projection attention that creates an anomaly detection map. We use this map as fuzzy guidance to our conditioned diffusion and show how it can autonomously correct facial images. We note that our application is trained on faces, and a flexible learning for scale adjustment would be needed to extend to other domains. Future work can also explore a frequency/latent-domain version of our method, for non-spatially-defined conditioning. 


\noindent \textbf{Acknowledgement:} Align Technology, Ringier, TX Group, NZZ, SRG, VSM, Viscom, and the ETH Zurich Foundation.

\newpage
\bibliographystyle{IEEEbib}
\bibliography{strings,refs}

\end{document}